\documentclass[12pt]{article}
\usepackage{amssymb,amsmath,cite,bm}
\usepackage[dvips]{graphicx,color}
\usepackage{psfrag,subfigure}
\newcommand{\norm}[1]{\left\Vert#1\right\Vert}

\newcommand{\diag}[1]{{\rm diag}\left(#1 \right)}

\newcommand{\col}[1]{{\rm col}\left(#1 \right)}

\newtheorem{theorem}{Theorem}
\newtheorem{remark}{Remark}

\title{Impedance Control for Manipulators Handling Heavy Payloads}
\author{Farhad Aghili \thanks{email: farhad.aghili@concordia.ca}}

\begin{document}

\date{}
\maketitle

\begin{abstract}
Attaching a heavy payload to the wrist force/moment (F/M) sensor of a manipulator can cause conventional impedance controllers to fail in establishing the desired impedance due to the presence of non-contact forces—namely, the inertial and gravitational forces of the payload. This paper presents an impedance control scheme designed to accurately shape the force-response of such a manipulator without requiring acceleration measurements. As a result, neither wrist accelerometers nor dynamic estimators for compensating inertial load forces are necessary. The proposed controller employs an inner-outer loop feedback structure, which not only addresses uncertainties in the robot’s dynamics but also enables the specification of a general target impedance model, including nonlinear models. Stability and convergence of the controller are analytically proven, with results showing that the control input remains bounded as long as the desired inertia differs from the payload inertia. Experimental results confirm that the proposed impedance controller effectively shapes the impedance of a manipulator carrying a heavy load according to the desired impedance model.
\end{abstract}

\section{Introduction}
\label{sec_introduction}

Impedance control, first proposed by Hogan \cite{Hogan-1985}, is used to shape the force-response of a manipulator interacting with its environment. In this approach, the contact force, measured by a wrist force/torque sensor, is utilized by the controller to establish a relationship between force and velocity based on the impedance model. The stability of conventional impedance controllers has been extensively analyzed \cite{Hogan-1987,Aghili-2010,Colgate-Hogan-1989,Aghili-Nmavar-2006}. Lasky {\em et al.} \cite{Lasky-Hsia-1991} introduced an inner/outer loop control scheme to compensate for uncertainties in robot dynamics using a position control algorithm in the inner loop. Adaptive impedance control schemes for robots with uncertain dynamic model parameters were later presented in \cite{Lu-Meng-1991,Carelli-Kelly-1991}. The stability of impedance control when dealing with unknown environments has been addressed in works such as \cite{Lee-Lee-1991,Aghili-Namvar-2004a,Jung-Hsia-Bonitz-2004,Namvar-Aghili-2003}. All of these approaches assume that direct measurements of external or contact forces/moments are available.

When a manipulator carries a heavy payload, conventional impedance control can fail to achieve the desired impedance. This failure occurs because the wrist force sensor measures not only external forces but also the inertial and gravitational forces of the payload \cite{Khatib-1987}. Additionally, a heavy payload can significantly alter the manipulator's dynamics. This is particularly relevant for space manipulators \cite{Aghili-2011k,Cyril-Misra-Ingham-Jaar-2000,Aghili-2009a,Aghili-Kuryllo-Okouneva-English-2010a,Aghili-Parsa-2007b,Aghili-2019e,Aghili-Parsa-2009b,Wee-Walker-1993,Aghili-2016c,Aghili-2022,Nenchev-Yoshida-1999,Aghili-2009d,Aghili-Parsa-2008a}, which are designed to handle heavy payloads by leveraging the weightlessness of space. For instance, the space manipulator aboard the International Space Station (ISS) can handle the Space Shuttle orbiter, which weighs approximately 100 metric tons, and employs impedance control for soft berthing with the ISS \cite{Aghili-Dupuis-Piedboeuf-deCarufel-1999}.

Approaches to address non-contact force components measured by a manipulator’s wrist F/M sensor typically focus on compensating for inertial force components using various estimation methods, rather than modifying the impedance control law itself \cite{Aghili-2005,Aghili-Piedboeuf-2003a}. Fujita and Inoue \cite{Fujita-Inoue-1979} explored inertial force estimation for cases with pre-planned trajectories. Uchiyama {\em et al.} \cite{Uchiyama-Kitagaki-1989} developed a method to estimate external forces and moments using an extended Kalman filter (EKF), though convergence of the EKF depends on the persistent excitation of input signals \cite{Kubus-Kroger-Wahl-2007,Aghili-p03,Kubus-Kroger-Wahl-2008}. Further methods, such as fusing force and accelerometer data via EKF, were proposed in \cite{Garcia-Robertsson-Ortega-Johansson-2006} and extended in \cite{Kubus-Kroger-Wahl-2008,Aghili-2012a} to estimate both inertial forces and the full set of inertial parameters of the load. However, these techniques require a 6-axis accelerometer. In contrast, our approach modifies the impedance control law itself, eliminating the need for dynamic estimators or accelerometers to account for inertial forces. This method avoids the challenges of input signal excitation and ensures system stability without requiring an additional sensor. Impedance control using inner/outer loops has been employed to emulate manipulators performing contact tasks \cite{Aghili-Piedboeuf-2002} and spacecraft dynamics \cite{Aghili-2005b,Aghili-Namvar-Vukovich-2006}. Frequency-domain analysis of impedance control for elastic-jointed industrial manipulators using an inner/outer loop configuration is discussed in \cite{Ferretti-Magnani-Rocco-2004}. Impedance control schemes for cooperative manipulators have also been explored, with efforts aimed at controlling object/environment interaction forces \cite{Schneider-Cannon-1992} and managing internal forces in dual-arm cooperative manipulators \cite{Caccavale-Chiacchio-Mario-Villani-2008}.

The goal of this paper is to present an impedance control scheme for manipulators carrying a heavy payload, without the need for compensating inertial force/moment measurements \cite{Aghili-2010}. We also investigate the limitations of such a scheme. The controller takes force/moment measurements from a sensor installed between the manipulator wrist and payload to establish the dynamic relationship between force error and position error based on a standard mass-damping-spring model. The impedance controller is further enhanced with an inner/outer loop feedback approach, allowing the specification of a general target impedance model. Using robust control theory \cite{Leitmann-1981,Aghili-2019c,Spong-Vidasagar-1989-Robust,Aghili-Buehler-Hollerbach-1997a}, we demonstrate that the control algorithm can overcome modeling uncertainties. Previous analyses of conventional impedance control show that contact stability imposes a lower bound on the ratio of the desired inertia matrix to the manipulator inertia matrix \cite{Whitney-1985,Bonitz-Hsia-1996,Aghili-Piedboeuf-2006}. However, we show that additional conditions on the payload, manipulator, and desired inertia matrices are required for the stability of impedance control when carrying a heavy payload \cite{Aghili-2010}.

This paper is organized as follows: Section \ref{sec:impedance_control} introduces the impedance control law for manipulators with heavy payloads. Section \ref{sec:nonlinear_impedance} further develops the impedance controller using an inner/outer loop control scheme, and Section \ref{sec:robust} presents a robust version of the controller to handle modeling uncertainties. Finally, Section \ref{sec:experiment} provides experimental results.

\section{Impedance Control}
\label{sec:impedance_control}
\begin{figure}[t]
\psfrag{Manipulator}[c][c][.9]{Manipulator}
\psfrag{Force/moment}[c][c][.9]{Force/moment}
\psfrag{sensor}[c][c][.9]{sensor}
\psfrag{Payload}[l][l][.9]{Payload} \psfrag{fext}[l][l][.9]{$\bm
f_{\rm ext}$} \psfrag{{W}}[c][c][.9]{$\{ {\cal W} \}$}
\psfrag{{T}}[c][c][.9]{$\{ {\cal C} \}$} \psfrag{{S}}[c][c][.9]{$\{
{\cal S} \}$} 
\centering \includegraphics[width=8.5cm]{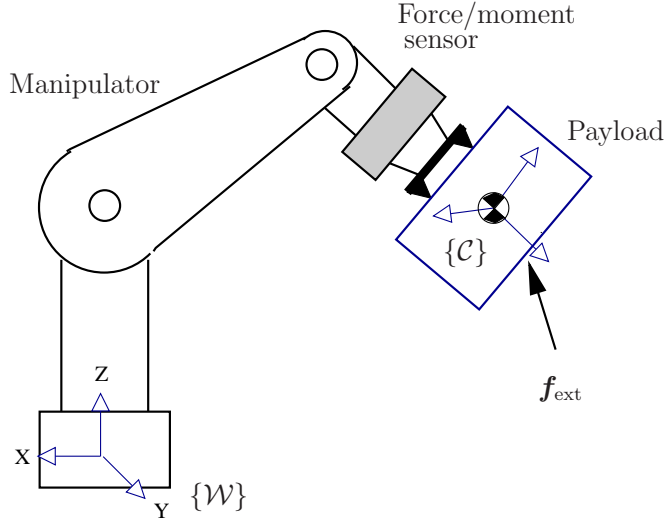}
\caption{A manipulator carrying a heavy
payload.}\label{fig:payload_robot}
\end{figure}

Fig.~\ref{fig:payload_robot} shows a manipulator carrying a payload
with non-negligible mass. We assume that no kinematic singularities
are encountered meaning that the robot can always operate in the
6-dimensional task space. A 6-axis force/moment sensor installed at
the manipulator wrist to measure the external force applied to the
payload. However due to the payload mass, the force sensor signal
$\bm f'_s$ contains the component of the generalized external force
${\bm f}'_{\rm ext}$ superimposed by the gravitational and inertial
forces of the payload. Let us assume that $\bm q$ is the vector of
the manipulator joint angles, vector $\bm x$ is the minimal
representation of the position and orientation of the payload, $\bm
v$ and $\bm \omega$ are the linear and angular velocities of the
payload expressed in the fixed-body frame $\{ {\cal C} \}$, and
$\bm\nu=\col{\bm v, \bm\omega}$ is the vector of {\em generalized
velocity}. Then, the following mapping through the manipulator
Jacobian matrix ${\bm J}'(\bm q)$ is in order
\begin{equation} \label{eq:Jacobian}
\bm\nu = \bm L(\bm q )\dot{\bm x} = {\bm J}'(\bm q) \dot{\bm q},
\end{equation}
where $\bm L(\bm q)= \diag{\bm 1_3 , \bm L_o(\bm q)}$ with $\bm 1_3$
being the $3\times3$ identity matrix, and the transformation matrix
$\bm L_o$ depends on a particular set of parameters used to
represent the orientation \cite{Canudas-Siciliano-Bastin-book-1996}.
The above kinematic mapping can also be written as  $\dot{\bm x} =
\bm J(\bm q) \dot{\bm q}$, where $\bm J(\bm q) = \bm L^{-1}(\bm q)
{\bm J}'(\bm q)$ is called the {\em analytical
Jacobian}~\cite{Canudas-Siciliano-Bastin-book-1996}; here we have
assumed that no {\em representation singularities} of the rotation
occurs. According to the {\em virtual work} principal, two sets of
generalized forces $\bm f'$ and $\bm f$ performing work on $\bm\nu$
and $\dot{\bm x}$ satisfy $\bm\nu^T \bm f'= \dot{\bm x}^T \bm f$,
and hence they are related by $\bm f_s= \bm L^T(\bm q) \bm f'_s$.
Similarly, $\bm f_{\rm ext}= \bm L^T(\bm q) \bm f'_{\bm ext}$.

Now, assume that the manipulator is cut right at its junction to the
payload. Note that the interaction force/moment between the two
separated systems of the manipulator and the payload are detected by
the force/moment sensor. Then, the equation of motion of the payload
in the task space can be written by
\begin{equation} \label{eq:payload_imp}
\bm M_p(\bm q) \ddot{\bm x}  + \bm h_p(\bm q, \dot{\bm q})  = - {\bm
f}_s + {\bm f}_{\rm ext},
\end{equation}
where
\begin{subequations}
\begin{align} \label{eq:Mp}
\bm M_p(\bm q) &= \begin{bmatrix}  m_p \bm 1_3 & \bm 0 \\ \bm 0 &  \bm L_o^T(\bm q) \bm I_p \bm L_o(\bm q) \end{bmatrix} \\
\label{eq:hp}
\bm h_p(\bm q, \dot{\bm q}) &= \begin{bmatrix} m_p \big( \bm\omega \times \bm v + g \bm R^T(\bm q) \bm k \big) \\
\bm L^T \big(\bm\omega \times \bm I_p \bm\omega \big) + \bm L^T(\bm
q) \bm I_p \dot{\bm L}_o (\bm q , \dot{\bm q}){\bm L}_o^{-1}(\bm q)
\bm\omega
\end{bmatrix}.
\end{align}
\end{subequations}
Here,  $m_p$ and $\bm I_p$ are the payload mass and inertia tensor,
rotation matrix $\bm R(\bm q)$ represents the end-effector attitude,
unit vector $\bm k$ is aligned with the gravity vector\footnote{If
the z-axis of the coordinate frame $\{W\}$ is perfectly parallel to
the earth's gravity vector, then $\bm k = \col{0,0,-1}$.} which is
expressed in the manipulator's base frame $\{ W \}$, and $g=9.81 \;
\mbox{m/s}^2$. Note that all force terms in the right-hand-side
(RHS) of \eqref{eq:payload_imp} are expressed in the fixed-body
coordinate frame $\{ C \}$ whose origin coincides with the payload's
center-of-mass. Note that the linear and angular velocities in
\eqref{eq:hp} can be computed from the manipulator's joint angles
and velocities using the kinematic relation \eqref{eq:Jacobian}, and
hence the nonlinear vector $\bm h_p$, which contains the Coriolis,
centrifugal and gravitational terms associated with the payload, can
be expressed as a function of the joint angles and velocities, i.e.,
$\bm h_p =\bm h_p(\bm q, \dot{\bm q})$.

It is known that the dynamics model of a manipulator in the task
space becomes \cite{Canudas-Siciliano-Bastin-book-1996}
\begin{equation} \label{eq:manipulator_imp}
\bm M_m(\bm q) \ddot{\bm x} + \bm h_m(\bm q, \dot{\bm q}) = \bm u +
\bm f_s,
\end{equation}
where
\begin{equation} \notag
\bm u = \bm J^{-T}(\bm q) \bm\tau,
\end{equation}
$\bm M_m(\bm q)$ is the Cartesian inertia matrix of the manipulator;
the nonlinear vector $\bm h_m(\bm q, \dot{\bm q})$ contains the
Coriolis, centrifugal and gravitational terms; and $\bm\tau$ is the
vector of joint torques (see Appendix~\ref{apdx:cartesian_dynamics}
for details). Assuming a kinematic singularity does not occur, there
is an one-to-one correspondence with the joint vector $\bm\tau$ and
auxiliary input $\bm u$. Therefore, in the following derivation, we
will take $\bm u$ as the control input for the sake of simplicity.
In the case of a redundant manipulator, the Jacobin inverse is not
unique. In solving the inverse kinematics of redundant manipulators,
there exist several joint torque optimization techniques in the
literature, one of which is the weighted pseudoinverse solution,
which can be used to maximize dynamic or static load capacity of the
manipulator \cite{Nakamura-1991}. Now, eliminating the interaction
force $\bm f_s$ from \eqref{eq:payload_imp} and
\eqref{eq:manipulator_imp}, we can express the combined dynamics of
the manipulator and the payload by
\begin{subequations}
\begin{equation} \label{eq:total_imp}
\bm M_t(\bm q) \ddot{\bm x} + \bm h_t(\bm q, \dot{\bm q}) = \bm u +
\bm f_{\rm ext},
\end{equation}
where
\begin{align} \label{eq:Mt_def}
\bm M_t(\bm q) &= \bm M_p(\bm q) + \bm M_m(\bm q),  \\
\bm h_t(\bm q, \dot{\bm q}) &= \bm h_p(\bm q, \dot{\bm q}) +  \bm
h_m(\bm q, \dot{\bm q}).
\end{align}
\end{subequations}
For the sake of notation simplicity, in the following developments
we consider all mass matrices and the associated nonlinear vectors
be functions of $\bm q$ and $\bm q, \dot{\bm q}$, respectively,
without writing their arguments unless otherwise is specified.

The desired impedance model that dynamically balances the external
contact force $\bm f_{\rm ext}$ is  typically chosen as the
second-order system \cite{Canudas-Siciliano-Bastin-book-1996}
\begin{equation} \label{eq:desired_imp}
\bm M_d (\ddot{\bm x} - \ddot{\bm x}_d) + \bm D_d(\dot{\bm x} -
\ddot{\bm x}_d) + \bm K_d (\bm x - \bm x_d) =\bm f_{\rm ext},
\end{equation}
where $\bm M_d$, $\bm D_d$ and $\bm K_d$ are the desired inertia,
damping and stiffness, respectively. Here, we assume that the
desired motion trajectory, $\bm x_d$, is specified to be twice
differentiable with respect to time. Eliminating the acceleration
from \eqref{eq:payload_imp} and \eqref{eq:desired_imp} and then from
\eqref{eq:total_imp} and \eqref{eq:desired_imp}, we get a set of two
equations:
\begin{subequations}
\begin{align} \label{eq:f_ext}
\bm f_{\rm ext} &= (\bm 1 - \bm M_p \bm M_d^{-1})^{-1} \big(\bm f_s + \bm h_p + \bm M_p \ddot{\bm x}_d \big)\\
\notag & + \big( \bm 1 - \bm M_d \bm M_p^{-1} \big)^{-1} (\bm D_d
\dot{\bm e} + \bm K_d \bm e), \\ \label{eq:control_law_fe} \bm u & =
\bm M_t \big( \ddot{\bm x}_d - \bm M_d^{-1} (\bm D_d \dot{\bm e} +
\bm K_d \bm e ) \big) + \bm h_t \\ \notag & + (\bm M_t \bm M_d^{-1}
- \bm 1 ) \bm f_{\rm ext}
\end{align}
\end{subequations}
where $\bm e=\bm x- \bm x_d$ is the position error. Finally,
substituting $\bm f_{\rm ext}$ from \eqref{eq:f_ext} into
\eqref{eq:control_law_fe} we obtain the overall impedance control
law as

\begin{subequations} \label{eq:imedance_control}
\begin{align} \notag
\bm u &= \big( \bm M_t(\bm q) - \bm\Gamma(\bm q) \bm M_p \big)
\big(\ddot {\bm x}_d - \bm M_d^{-1} (\bm D_d \dot{\bm e} + \bm K_d
\bm e) \big)
\\  \label{eq:control_law_fs} &+  \bm h_t(\dot{\bm q},\bm q)- \bm\Gamma(\bm q)  \big(
\bm f_s + \bm h_p(\dot{\bm q},\bm q) \big)
\end{align}
where
\begin{equation} \label{eq:Gamma}
\bm\Gamma(\bm q) \triangleq \bm 1 - \bm M_m \bm M_d^{-1} \big(\bm 1
- \bm M_p \bm M_d^{-1} \big)^{-1}.
\end{equation}
\end{subequations}
By inspection, one can show that the  impedance control law
\eqref{eq:imedance_control} coincides with the conventional
impedance control law \cite{Canudas-Siciliano-Bastin-book-1996} if
$\bm M_p \equiv \bm 0$.

From a stability point of view, the fundamental difference between
impedance control of manipulators  without and with a payload is
that, unlike the former case, the latter does not always lead to a
stable system. To this end, denote
\begin{equation} \label{eq:Delta_d}
\bm\Delta_d \triangleq \bm M_p \bm M_d^{-1}.
\end{equation}
Then, it is apparent from expressions \eqref{eq:control_law_fs} and
\eqref{eq:Gamma} that the control input remains bounded if matrix
$\bm 1- \bm\Delta_d$ is not singular, i.e.,
\begin{equation} \label{eq:det_alpha}
\det( \bm 1 - \bm\Delta_d) \neq 0.
\end{equation}
Since $\det(\bm 1 - \bm\Delta_d) = \prod_i(1 -
\lambda_i(\bm\Delta_d))$, we can say that \eqref{eq:det_alpha} is
equivalent to
\begin{equation} \label{eq:cond_bounded}
\lambda_i(\bm\Delta_d) \neq 1 \qquad \forall i =1, \cdots, n,
\end{equation}
where $\lambda_i(\cdot)$ denotes the $i$th eigenvalue of matrix
$(\cdot)$. In  the case that the desired inertia is a diagonal
matrix with all of its diagonal elements equal to $m_d$, i.e., $\bm
M_d = m_d \bm 1_n$, the above condition is reduced to
\begin{equation} \label{eq:scalar_cond}
m_d \neq \lambda_i(\bm M_p)  \qquad \forall i =1, \cdots, n .
\end{equation}

\begin{theorem}
Assume that the desired inertia matrix is selected such that
\eqref{eq:cond_bounded} is satisfied. Then, applying control law
\eqref{eq:control_law_fs} to the manipulator system
\eqref{eq:manipulator_imp} attached to a payload with generalized
inertia $\bm M_p$ establishes the desired impedance
\eqref{eq:desired_imp}.
\end{theorem}

\section{Impedance Control with Inner/Outer Loop}
\label{sec:nonlinear_impedance}
\subsection{Nonlinear Impedance Model}

In this section, we extend the impedance control of manipulators
carrying a heavy payload using an inner/outer loop control scheme in
order to enhance the system robustness with respect to the robot
dynamics uncertainties. We also assume that the desired impedance
model relating the force and motion  takes the following general
form
\begin{equation} \label{eq:desired_nonlin}
\bm M_d \ddot{\bm x} + \bm h_d(\dot{\bm x}, \bm x) = \bm f_{\rm
ext}.
\end{equation}
It is worthwhile mentioning that an interesting application of such
a impedance control approach is in zero-g emulation of a scaled
spacecraft prototype  under the test in a 1-g laboratory environment
\cite{Aghili-Namvar-Vukovich-2006,Aghili-Namvar-2008}.

Let us define an estimation of the acceleration $\ddot{\bm
x}^{\star}$ obtained by substituting $\bm f_{\rm ext}$ from
\eqref{eq:payload_imp} into \eqref{eq:desired_nonlin}, i.e.,
\begin{subequations} \label{eq:outer}
\begin{equation} \label{eq_DeltaDynamics}
\bm M_{\Delta} \ddot{\bm x} ^{\star} + \bm h_{\Delta}(\dot{\bm x},
\bm x) = {\bm f}_s,
\end{equation}
where
\begin{align*} \label{eq_M_Delta}
\bm M_{\Delta} \triangleq  \bm M_d - \bm M_p \quad \mbox{and} \quad
\bm h_{\Delta} \triangleq \bm h_d - \bm h_p.
\end{align*}
It should be pointed out that in derivation of
\eqref{eq_DeltaDynamics}, we presumed that the manipulator's
acceleration would follow the acceleration trajectory according to
the impedance model \eqref{eq:desired_nonlin}. We will prove that
this is case later. Therefore, the acceleration estimate, $\ddot{\bm
x}^*$, and actual acceleration, $\ddot{\bm x}$, are not considered
identical. In other words, as will be shown later in this section,
the manipulator establishes the desired impedance
\eqref{eq:desired_nonlin} only if the actual acceleration $\ddot{\bm
x}$ follows $\ddot{\bm x} ^{\star}$.  In the following development,
we will show that the difference between the two accelerations
converges to zero under an adequate control law.
\begin{remark}
By inspection, one can show that if condition
\eqref{eq:cond_bounded} is satisfied, then matrix $\bm M_{\Delta}$
is invertible.
\end{remark}
Therefore, if \eqref{eq:cond_bounded} is satisfied, then the
velocity estimate, $\dot{\bm x}^{\star}$, can be obtained through a
numerical integration of \eqref{eq_DeltaDynamics}, i.e.,
\begin{equation} \label{eq:dotxstar}
\dot{\bm x}^{\star}(t) = \int_0^t \bm M_{\Delta}^{-1} \big( {\bm
f}_s - \bm h_{\Delta} \big) {\rm d} \tau.
\end{equation}
\end{subequations}
Similarly, ${\bm x}^{\star}$ can be obtained by the second
integration. Note that $\ddot{\bm x}^{\star}$ and $\ddot{\bm x}$ are
not necessarily equal and neither do their integrated values.

Now, the objective is to force the manipulator to follow the
trajectory dictated by \eqref{eq:dotxstar}. It seems that this goal
can be achieved by an inverse dynamics controller
\cite{Spong-Vidasagar-1989,Canudas-Siciliano-Bastin-book-1996}
designed for system \eqref{eq:manipulator_imp}. However, such a
control design scheme will lead to an algebraic loop, which is not
realizable in physical systems unless a time-delay is introduced
into the control system. Note that the force sensor signal $\bm f_s$
contains components of the inertial forces of the payload due to the
acceleration. Thus, compensating for the perturbation $\bm f_s$
results in a torque control law which has a direct component of the
acceleration, while the acceleration itself is algebraically related
to the joint torques. This problem can be avoided by using an
inverse-dynamics controller based on the complete model of the
manipulator and payload \eqref{eq:total_imp} (note that the payload
and manipulator together forms a single multibody chain system) that
requires compensating only for the perturbation ${\bm f}_{\rm ext}$,
which is not correlated to the inertial forces. However, the
difficulty in this approach is that because of the acceleration
estimation error, one can only obtain an estimation of the external
force ${\bm f}_{\rm ext}$; as will be show in the following
analysis. Let
\begin{equation} \label{eq:delddotx}
\ddot {\tilde {\bm x}}\triangleq \ddot {\bm x}^{\star} - \ddot{\bm
x}
\end{equation}
denotes the acceleration error. Now, upon substitution of $\ddot
{\bm x}^{\star}$ from \eqref{eq_DeltaDynamics} into
\eqref{eq:delddotx} and then substituting the resultant acceleration
term into \eqref{eq:payload_imp}, we can write the expression of the
external force as:
\[ {\bm f}_{ext} =  {\bm f}_{\rm ext}^{\star} + \tilde {\bm f}_{\rm ext},\]
where
\begin{equation} \label{eq:fstar}
{\bm f}_{\rm ext}^{\star} = \big( \bm 1 + \bm M_p \bm
M_{\Delta}^{-1} \big){\bm f}_s + \bm h_p - \bm M_p \bm
M_{\Delta}^{-1} \bm h_{\Delta}
\end{equation}
is the external force estimate and
\begin{equation} \label{eq:tilde_F}
\tilde {\bm f}_{ext} = - \bm M_p \ddot {\tilde {\bm x}}
\end{equation}
is the corresponding estimation error. Clearly, the force estimation
error goes to zero if and only if the acceleration error does so.

Now, considering the force estimation \eqref{eq:fstar} for
compensating the force perturbation, we propose the following inner
loop control
\begin{align} \notag
\bm u & = \bm M_t(\bm q) \ddot{\bm x}^{\star} + \bm h_t - {\bm f}^{\star}_{\rm ext} \\
\label{eq:u_law} & + \bm M_t(\bm q) \big( \bm G_d (\dot{\bm
x}^{\star} - \dot{\bm x} ) + \bm G_p ( \bm x^{\star} - \bm x )
\big),
\end{align}
where $\ddot{\bm x}^{\star}$, $\dot{\bm x}^{\star}$,  ${\bm
x}^{\star}$ and $\bm f_{\rm ext}^{\star}$ are obtained from
\eqref{eq:desired_nonlin}- \eqref{eq:fstar}, while $\bm G_d>0$ and
$\bm G_p >0$ are the feedback gains. In the following analysis, we
will show that the inner loop controller \eqref{eq:u_law} in
conjunction with the outer loop \eqref{eq:outer} can shape the
impedance of the combined system of manipulator and payload
according to \eqref{eq:desired_nonlin} provided that a condition on
the system mass distribution is met. Substituting \eqref{eq:u_law}
into the dynamics model \eqref{eq:total_imp} yields the equation of
force error as
\begin{equation} \label{eq:Mterror}
\bm M_t \big( \ddot {\tilde{\bm x}} + \bm G_d \dot{\tilde{\bm x}} +
\bm G_p {\tilde{\bm x}} \big) = - \tilde {\bm f}_{\rm ext}.
\end{equation}
Using \eqref{eq:tilde_F} in \eqref{eq:Mterror} yields the following
autonomous system
\begin{align} \notag
\big( \bm 1 - \bm M_t^{-1} \bm M_p \big) \ddot {\tilde{\bm x}} + \bm
G_d \dot{\tilde{\bm x}} + \bm G_p {\tilde{\bm x}} &= \bm 0\\
\label{eq:autonomous} \big( \bm 1 - \bm M_t^{-1} \bm M_p \big) \big(
\ddot {\tilde{\bm x}} + \bm G_d \dot{\tilde{\bm x}} + \bm G_p
{\tilde{\bm x}} \big) + \bm M_t^{-1} \bm M_p \big( \bm G_d
\dot{\tilde{\bm x}} + \bm G_p {\tilde{\bm x}} \big) &= \bm 0
\end{align}
Multiplying both sides of \eqref{eq:autonomous} by $\bm M_m^{-1} \bm
M_t$ and using identity \eqref{eq:Mt_def} in the resultant equations
yields
\begin{equation} \label{eq_accODE} \ddot
{\tilde{\bm x}} +  \bm G_d \dot{\tilde{\bm x}} + \bm G_p {\tilde{\bm
x}} + \bm\Delta_m \big( \bm G_d \dot{\tilde{\bm x}} + \bm G_p
{\tilde{\bm x}} \big)= \bm 0,
\end{equation}
where
\begin{equation} \label{eq:Delta_m}
\bm\Delta_m \triangleq \bm M_m^{-1} \bm M_p.
\end{equation}
Stability of the closed-loop system \eqref{eq_accODE} remains to be
proved. We will show in the following that system \eqref{eq_accODE}
remains stable if the coefficient matrix of the additive term, i.e.,
$\bm\Delta_m$, is sufficiently small. Let us assume that $\bm z
=\mbox{col}( \tilde{\bm x}, \dot{\tilde {\bm x}})$ represents the
state vector. Then, \eqref{eq_accODE} can be written in the
following compact form
\begin{equation} \label{eq:perturbed}
\dot{\bm z} = \bm A \bm z + \bm d(t, \bm z),
\end{equation}
where
\[ \bm A=\begin{bmatrix} \bm 0 & \bm 1 \\ -\bm G_p & - \bm G_d \end{bmatrix} \quad \text{and}
\quad \bm d (t, \bm z)=   -\bm D \bm\Delta_m \bm G \bm z,\] with
$\bm G=[\bm G_p \quad \bm G_d]$ and $\bm D^T=[\bm 0 \quad \bm 1]$.
The perturbation term $\bm d$ satisfies the linear growth bound
\[ \| \bm d \| \leq  \| \bm G \|   \|\bm\Delta_m \| \| \bm z \|, \]
where
\[\frac{\lambda_{\rm min}(\bm M_p)}{\lambda_{\rm max}(\bm
M_m)} \leq \norm{\bm\Delta_m} \leq \frac{\lambda_{\rm max}(\bm
M_p)}{\lambda_{\rm min}(\bm M_m)} < \infty \qquad \forall \bm z \in
\mathbb{R}^{2n}.\] Therefore, system \eqref{eq:perturbed} is in the
form of {\em vanishing perturbation} \cite{Khalil-1992}. Here,
$\norm{\cdot}$ denotes the Euclidean norm of a vector or a matrix.
Moreover, since $\bm A$ is Hurwitz, one can show that the perturbed
system can be globally exponentially stable if the gains are
adequately selected, i.e.,
\begin{equation} \label{eq:Deltam<alpha}
\norm{\bm\Delta_m} \leq \kappa(\bm G),
\end{equation}
where $\kappa(\bm G)$ is a function of the feedback gains; see the
Appendix~\ref{apdx:Lyapunov} for details. That means there must
exist scalar $\mu
>0$ such that $\| \bm z \| \leq \norm{\bm z(0)} e^{-\mu t}$.
Therefore, it can be inferred from \eqref{eq_accODE} that
\begin{equation} \label{eq_bound_ddq}
\| \ddot{\tilde {\bm x}} \| \leq \phi e^{-\mu t},
\end{equation}
where scaler $\phi$ depend on the initial error
\[ \phi = (1 +\norm{\bm\Delta_m}) \norm{\bm G } \bm z (0) . \] Now,
we are ready to derive the input/output relation of the closed loop
system under the proposed control law. Adding both sides of
\eqref{eq:payload_imp} and \eqref{eq_DeltaDynamics} yields
\begin{subequations} \label{eq:purturbed_delta}
\begin{equation} \label{eq_purturbed}
\bm M_d \ddot{\bm x} + \bm h_d(\dot{\bm x}, \bm x) = {\bm f}_{\rm
ext} + \bm\delta,
\end{equation}
where
\begin{equation} \label{eq_delta}
\bm\delta(t)= -\bm M_{\Delta} \ddot{\tilde {\bm x}}.
\end{equation}
\end{subequations}
It follows from \eqref{eq_bound_ddq}  and \eqref{eq_delta} that
\begin{equation} \label{eq_del}
\| \bm\delta \| \leq  \phi \lambda_{\rm max}({\bm M}_{\Delta})
e^{-\mu t},
\end{equation}
which means that the perturbation $\bm\delta$ exponentially relaxes
to zero from its initial value.

To summarize, consider the target impedance dynamics
\eqref{eq:desired_nonlin} for a manipulator with inertia $\bm M_m$
carrying a payload with $\bm M_p$. Assume that the inertia ratios
$\norm{\bm\Delta_d}$ and $\norm{\bm\Delta_m}$ satisfy conditions
\eqref{eq:cond_bounded} and \eqref{eq:Deltam<alpha}, respectively.
Then, the force-response of the combined manipulator and payload
under the impedance controller with the inner/outer loops
\eqref{eq:outer}-\eqref{eq:u_law} converges to the target impedance
\eqref{eq:desired_nonlin}, while the control input remains bounded.

It should be pointed out that in addition to \eqref{eq:cond_bounded}
and \eqref{eq:Deltam<alpha}, the condition for contact stability
must be also satisfied. The problem of coupled or contact stability
has been thoroughly examined in the literature
\cite{Whitney-1985,Eppinger-Seering-1986,Kazerooni-1987,Lawrence-1988,Colgate-Hogan-1989,Colgate-1994,Bonitz-Hsia-1996,Aghili-Piedboeuf-2006}.
Coupled stability of a feedback controller for a manipulator
interacting with passive environments was studied in
\cite{Colgate-1994}, while the instability attributed to
noncolocation of the sensor and actuators was investigated in
\cite{Eppinger-Seering-1986,Colgate-Hogan-1989}. The effect of
delays in contact stability of impedance control is analyzed in
\cite{Lawrence-1988}. Digital implementations of impedance control
and contact stability were examined in
\cite{Whitney-1985,Bonitz-Hsia-1996}, while stability analysis of
impendence control under the presence of actuator dynamics was
carried out in \cite{Aghili-Piedboeuf-2006}. The analysis results of
\cite{Whitney-1985,Bonitz-Hsia-1996,Aghili-Piedboeuf-2006} showed
that contact stability imposes a lower bound limit on the magnitude
of the ratio of the desired inertia matrix to the actual effective
inertia matrix when the impedance control is implemented digitally
or the actuators are of finite bandwidth.

\subsection{Robust Impedance Control} \label{sec:robust}
It is much more reasonable to suppose that the model-based control
law \eqref{eq:u_law} is actually computed from $\hat{\bm M}_m$ and
$\hat{\bm h}_m$, which are nominal or computed versions of $\bm M_m$
and $\bm h_m$ \cite{Spong-Vidasagar-1989-Robust}. In that case, the
perturbation in system \eqref{eq:perturbed} will be compounded by an
additional term due to the modeling uncertainty. In this section,
the control law will be redesigned  in order to overcome the effects
of modeling uncertainties and the perturbation all together. To this
end, the control law \eqref{eq:u_law} is modified to
\begin{equation}
\bm u = \hat{\bm u} +  \hat{\bm M}_t  \bm w,
\end{equation}
where $\hat{\bm u}$ is control law \eqref{eq:u_law} computed from
the nominal variables $\hat{\bm M}_m$ and $\hat{\bm h}_m$, while
$\bm w$ is the additional control input to compensate for the
dynamics uncertainty and perturbation \cite{Leitmann-1981}.
Substituting the above control law into \eqref{eq:total_imp} yields
\begin{align} \label{eq:ddot_tilde}
\ddot{\tilde {\bm x}} = \tilde{\bm M}_t \ddot{\bm x}^{\star}  + \bm
M_t^{-1} (\tilde{\bm f}_{\rm ext} -   \tilde{\bm h}_t) - \bm
M_t^{-1} \hat{\bm M}_t( \bm w + \bm G \bm z ),
\end{align}
where $\tilde{\bm h}_t=  \tilde{\bm h}_m=\bm h_m - \hat{\bm h}_m$
and $\tilde{\bm M}_t= \bm 1 - \bm M_t^{-1} \hat{\bm M}_t$ represents
the parametric uncertainty or modeling error. On the other hand,
using identity \eqref{eq:delddotx} in \eqref{eq_delta}, one can
relate the desired acceleration $\ddot{\bm x}^{\star}$ to the
external force by
\begin{equation} \label{eq:ddotx_star}
\ddot{\bm x}^{\star} = \bm M_d^{-1} \big( \bm f_{\rm ext}
  - \bm h_d \big) + \bm\Delta_d^T \ddot{\tilde {\bm x}}.
\end{equation}
Substituting $\ddot{\bm x}^{\star}$ and $\tilde{\bm f}_{\rm ext}$
from \eqref{eq:ddotx_star} and  \eqref{eq:tilde_F} into
\eqref{eq:ddot_tilde} yields the following first-order matrix
differential equation
\begin{equation}
\dot{\bm z} = \bm A \bm z + \bm D(\bm\eta - \bm w),
\end{equation}
where $\bm z$ has been already defined as the state vector and
\begin{subequations}
\begin{align}\notag
\bm\eta & =\bm\Delta \tilde{\bm M}_t \bm M_d^{-1}(\bm f_{\rm ext} -
\bm h_d)  - \bm\Delta \bm M_t^{-1} \tilde{\bm h}_t \\
\label{eq:eta}
 & + (\bm 1 - \bm\Delta \bm M_t^{-1} \hat{\bm M}_t)
(\bm G \bm z + \bm w), \\ \label{eq:Delta}  \bm\Delta &= \big( \bm 1
+ \bm\Delta_t^T - \tilde{\bm M}_t \bm\Delta_d^T \big)^{-1},\\
\bm\Delta_t& =\bm M_p \bm M_t^{-1}.
\end{align}
\end{subequations}
The control input $\bm w$ is chosen according to robust control
theory \cite{Leitmann-1981,Spong-Vidasagar-1989-Robust}:
\begin{equation} \label{eq:switch_controller}
\bm w = \left\{ \begin{array}{lll} - \frac{\bm D^T \bm P \bm
z}{\|\bm D^T \bm P \bm z \|} \rho & \quad & \mbox{if} \quad
\norm{\bm D^T \bm P \bm z}
\neq 0 \\
\bm 0 & \quad & \mbox{otherwise}
\end{array} \right.
\end{equation}
where $\rho$ should be chosen so that
\begin{equation} \label{eq:rho_eta}
\rho \geq \norm{\bm\eta} \qquad \forall \bm x, \dot{\bm x}.
\end{equation}
In view of definition of $\bm\eta$ in \eqref{eq:eta} and that
$\norm{\bm w}=\rho$, assumption \eqref{eq:rho_eta} makes sense only
if
\begin{equation} \label{eq:alpha}
\| \bm 1 - \bm\Delta \bm M_t^{-1} \hat{\bm M}_t \| \leq \alpha <1.
\end{equation}
Note that \eqref{eq:alpha} automatically implies that the matrix
inversion in \eqref{eq:Delta} is well behaved. The conventional
robust position control of manipulators assume that $\| \tilde{\bm
M}_t \| \leq 1$ and $\| \tilde {\bm h}_t \| < \infty\quad\forall \bm
x, \dot{\bm x}$ \cite{Spong-Vidasagar-1989-Robust}. In addition to
the later assumption, we need to assume that  $\|\bm f_{\rm ext} \|$
and $\| \bm h_d \|$ are bounded quantities too. Note that the
nonlinear term $\bm h_d$ is a part of the desired impedance model,
eq.(13), which is a mathematical definition rather than describing a
physical phenomenon. Therefore, the desired nonlinear term $\bm h_d$
can be easily  defined to be a bounded quantity (e.g., using
saturation functions). Under this circumstances, condition
\eqref{eq:rho_eta} can be always satisfied (see
Appendix~\ref{apdx:rho}). Then, a standard argument shows that the
error $\bm z$ asymptotically converges to zero under the robust
control feedback. To this end, selecting the Lyapunov function
\eqref{eq:Lyapunov} again, we can show that its time-derivative
satisfies
\begin{equation}
\dot{V} \leq - \norm{\bm z}^2 + 2 \| \bm D^T \bm P \bm z \| \big(
\norm{\bm\eta} - \rho \big) <0.
\end{equation}
Assumption \eqref{eq:alpha} is more stringent than the peer
assumption made  in the conventional robust position control of
manipulators \cite{Spong-Vidasagar-1989-Robust}, i.e., $\|
\tilde{\bm M}_t \| \leq \alpha < 1$. It is worth mentioning that if
the payload inertia is negligible, i.e., $\bm M_p \approx 0
\Rightarrow \bm\Delta \approx \bm 1$, then \eqref{eq:alpha}
coincides with the assumption of the conventional robust position
control of manipulators.

It is known that the discontinuity of the additional control input
\eqref{eq:switch_controller} together with the finite bandwidth of
sensors and actuators employed in the control system can lead to
undesirable {\em chattering} \cite{Spong-Vidasagar-1989-Robust}. In
that case, the trajectories of the state vector oscillates around
the sliding hyperplane $\bm D^T \bm P \bm z= \bm 0$ at high
frequency instead of being attracted to the hyperplane and then
moving toward the origin. It is possible to eliminate the chattering
by allowing the error $\bm z$ to vary within a boundary layer whose
thickness is chosen as $\varepsilon$. This kind of controller is
discussed in \cite{Leitmann-1981}
\begin{equation} \label{eq:chatter_free}
\bm w = \left\{ \begin{array}{lll} - \frac{\bm D^T \bm P \bm
z}{\|\bm D^T \bm P \bm z \|} \rho & \quad & \mbox{if} \quad \norm{\bm D^T \bm P \bm z}\leq \varepsilon \\
\bm 0 & \quad & \mbox{otherwise}
\end{array} \right.
\end{equation}

\section{Experiment}
\label{sec:experiment}

\begin{table}
\caption{Manipulator's link parameters.} \centering{
\begin{tabular}{lllllllll}
\hline\hline
Link & m {\scriptsize(kg)}& $I_{xx}$ \scriptsize{(kgm$^2$)} & $I_{yy}$ \scriptsize{(kgm$^2$)}& $I_{zz}$ \scriptsize{(kgm$^2$)} & $c_{x}$ {\scriptsize(m)}& $c_{y}$ {\scriptsize(m)} & $c_{z}$ {\scriptsize(m)}   \\
\hline
1 & 27.31 & 0.31 & 0.38 & 0.30 & -0.140 & -0.044 & 0.170\\
2 & 21.00 & 0.25 & 3.35 & 3.30 & -0.490 & 0.000 & 0.125\\
3 & 10.00 &  0.17 & 2.60 & 2.50 & -0.401 & 0.000 & 0.030\\
4 & 4.33 & 0.035 & 0.05 & 0.02 & -0.166 & 0.079 & 0.171\\
5 & 4.02 & 0.026 & 0.04 & 0.02 & -0.162 & -0.009 & 0.204\\
6 & 1.59 & 0.01 & 0.01 & 0.003  & 0.000 & 0.000 & 0.220 \\
\hline\hline
\end{tabular}} \label{tab:link_param}
\end{table}

This section describes experimental results obtained from
implementation of the proposed impedance control scheme using a
robotic manipulator described in \cite{Aghili-2010}. The inertial
properties of the manipulator links have been identified
\cite{Lamarche-2002} as listed in Table~\ref{tab:link_param}. The
manipulator joints are equipped with torque sensors which are used
by an inner loop feedback controller to minimize the joint friction.

A dummy box  is mounted on the manipulator wrist. The payload mass
and inertia are:
\begin{equation}\notag
m_p=16~\text{kg} \quad  \text{and} \quad \bm I_p=\begin{bmatrix} 0.33 & 0 & 0 \\
0 & 0.62 & 0 \\ 0 & 0 & 0.71 \end{bmatrix}~\text{kgm}^2 .
\end{equation}
The Cartesian inertia matrix of the manipulator calculated at a
nominal position within the workspace used for the tests is:
\begin{equation} \notag
\bm M_m = \begin{bmatrix}
57.73 &  13.53 &  -3.34 &  -0.56  & -6.21  & 18.05\\
13.53 &  69.26 & -19.83 &  -1.40  & -5.52  & 18.23\\
-3.34 & -19.83 &  38.89 &  -4.48  &  5.88  & -9.04\\
-0.56 & -1.40  & -4.48  & 13.23   & -0.75  &  0.22\\
-6.21 &  -5.52 &   5.88 &  -0.75  & 13.30  & -8.68\\
18.05 &  18.23 &  -9.04 &   0.22  & -8.68  & 18.26
\end{bmatrix}
\end{equation}

The force/moment interaction between the manipulator and the payload
were measured by a six-axis JR3 force/moment sensor. The impedance
controller scheme with inner/outer loop described in
Section~\ref{sec:nonlinear_impedance} was developed and implemented
using Simulink. Matrix manipulation was performed by using the DSP
Blockset of Matlab/Simulink \cite{Matlab_DSP}. The Real-Time
Workshop package \cite{Matlab_RTW} generated portable C code from
the Simulink model which was executed on a QNX real-time operating
system.

\begin{figure}
\psfrag{force}[c][c][.8]{force~(N)}
\psfrag{torque}[c][c][.8]{moment~(Nm)}
\psfrag{fx}[l][l][.8]{$f_x$}\psfrag{fy}[l][l][.8]{$f_y$}\psfrag{fz}[l][l][.8]{$f_z$}
\psfrag{tx}[l][l][.8]{$n_x$}\psfrag{ty}[l][l][.8]{$n_y$}\psfrag{tz}[l][l][.8]{$n_z$}
\centering
\includegraphics[width=9.5cm]{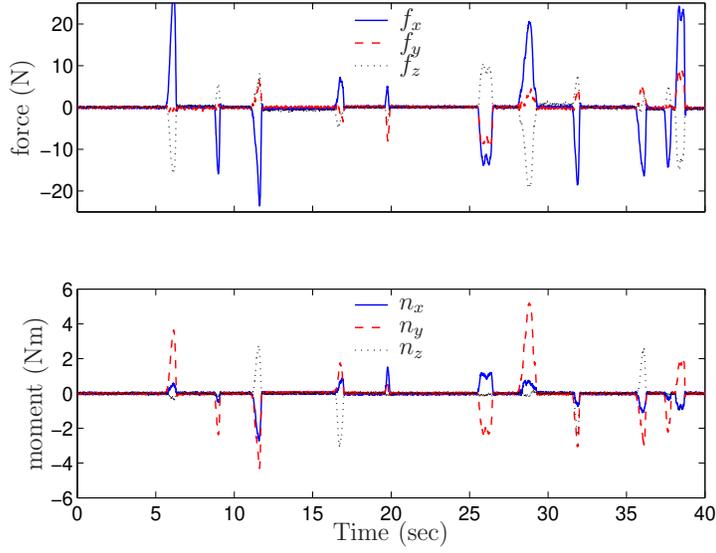}
\caption{Dynamics force and moment.}\label{fig:force}
\end{figure}

\begin{figure}
\psfrag{vx}[l][l][.8]{$v_x$}\psfrag{vy}[l][l][.8]{$v_y$}\psfrag{vz}[l][l][.8]{$v_z$}
\psfrag{wx}[l][l][.8]{$\omega_x$}\psfrag{wy}[l][l][.8]{$\omega_y$}\psfrag{wz}[l][l][.8]{$\omega_z$}
\psfrag{vsim}[c][c][.8]{$\bm v_{\rm
sim}$~(m/s)}\psfrag{omegsim}[c][c][.8]{$\bm\omega_{\rm
sim}$~(rad/s)} \psfrag{v}[c][c][.8]{$\bm
v$~(m/s)}\psfrag{omeg}[c][c][.8]{$\bm\omega$~(rad/s)} \centering{
\subfigure[Actual
trajectories]{\includegraphics[clip,width=9.5cm]{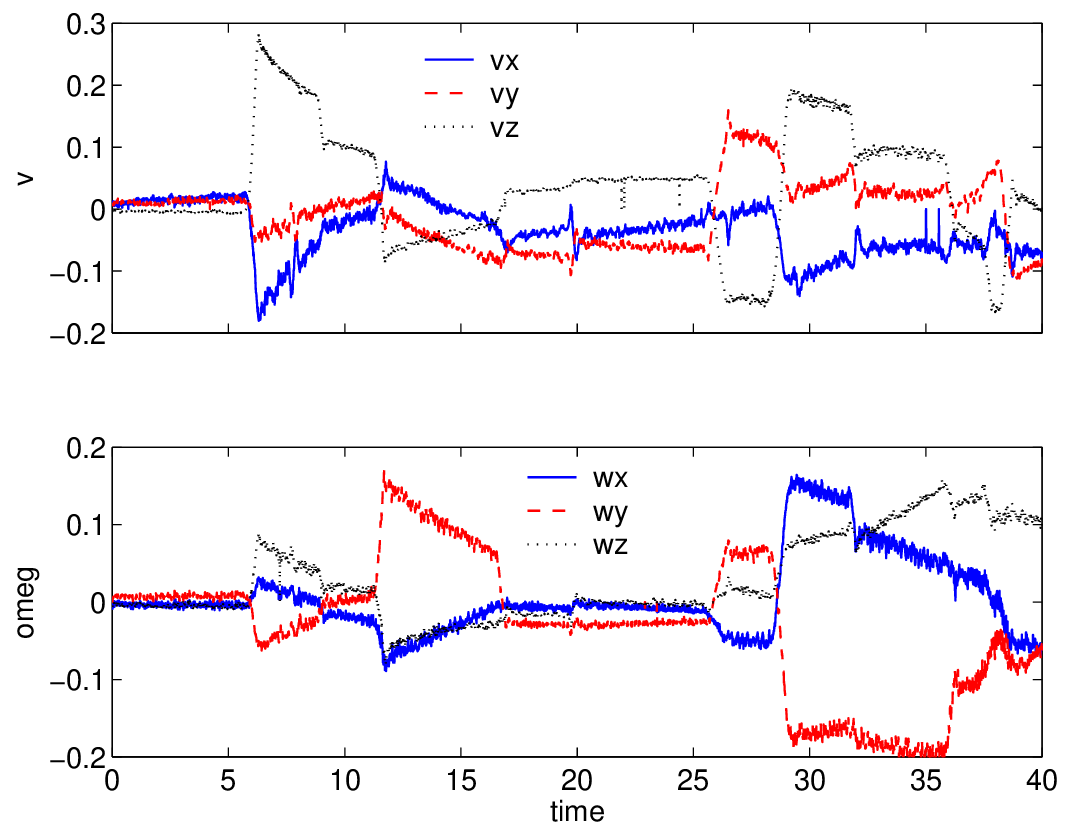}\label{fig:vel_act}}\\
\subfigure[Simulated
trajectories]{\includegraphics[clip,width=9.5cm]{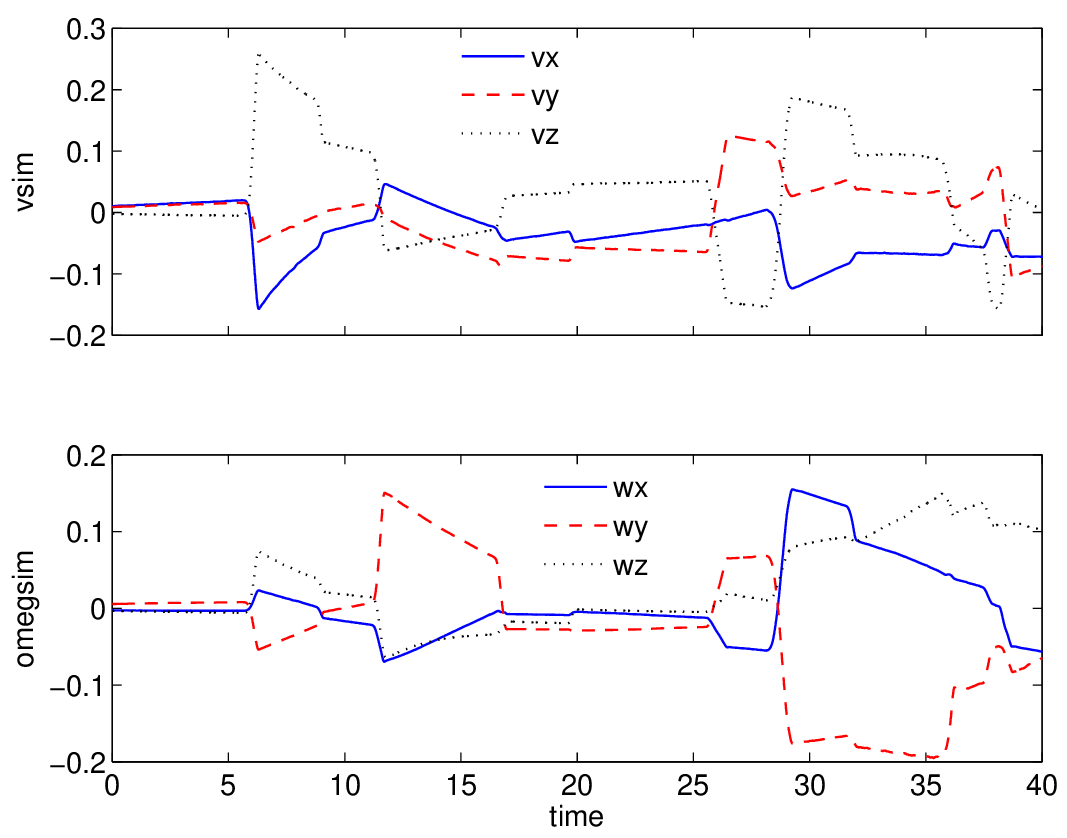}\label{fig:vel_sim}}
\caption{Trajectories of the linear and angular
velocities.}\label{fig:velocity}}
\end{figure}

The objective of the first case study is to show that the proposed
controller is able to match the manipulator force-response according
to the desired impedance even though the force sensor signal is
directly incorporated into the controller without compensating for
the inertial forces. To this end, the target mass and inertia of the
impedance controller is set to be three times as much as the actual
mass and inertia of the payload, i.e.,
\begin{equation} \label{eq:target_inertia}
\bm M_d = 3 \bm M_P.
\end{equation}
Fig.~\ref{fig:force} shows trajectories of the forces and moments
due to external force impulses applied by hand. These trajectories
are obtained by compensating for the gravitational and inertial
force components of the force/moment sensor output through an
off-line process; see Appendix~\ref{apdx:external_force} for the
derivation of the external force as a function of $\bm f_s$ and $\bm
u$. The subsequent trajectories of the linear and angular velocities
of the payload are illustrated in Fig.~\ref{fig:vel_act}. The
simulated velocity profiles in response to the same force/moment
inputs according to the target impedance model are also depicted in
Fig.~\ref{fig:vel_sim}. A comparison between trajectories of
Figs.~\ref{fig:vel_act} and \ref{fig:vel_sim} shows that the
impedance controller has succeeded to establish the target impedance
characteristic even though the wrist force sensor sees inertial
forces of the load. In order to quantify the difference between the
simulated velocity and the robot velocity, the root mean square
error (RMSE) of the linear velocity is given in percentage as:
\[ {\rm RMSE}(\bm v) = 100 \times \sqrt{ \frac{\int_0^T \| \bm v - \bm v_{\rm sim} \|^2
dt}{\int_0^T \| \bm v\|^2 dt}} = 6.1 \%. \] Similarly, the RMSE of
the angular velocity is calculated to be
\[ {\rm RMSE}(\bm\omega) = 4.3 \%. \]

\psfrag{time}[c][c][.8]{Time~(sec)}

\begin{figure}
\psfrag{xd}[l][l][.8]{$x_{d_z}$} \psfrag{x}[l][l][.8]{$x_z$}
\psfrag{position}[c][c][.8]{position~(m)} \centering
\includegraphics[width=9.5cm]{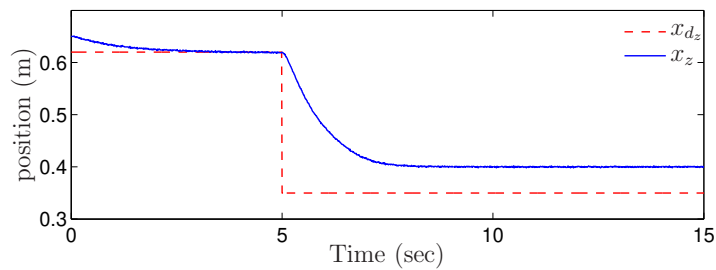}
\caption{The $z$-axis position of the
payload.}\label{fig:contact_pos}
\end{figure}

\begin{figure}
\psfrag{force}[c][c][.8]{force~(N)}
\psfrag{torque}[c][c][.8]{moment~(Nm)}
\psfrag{fx}[l][l][.8]{$f_x$}\psfrag{fy}[l][l][.8]{$f_y$}\psfrag{fz}[l][l][.8]{$f_z$}
\psfrag{tx}[l][l][.8]{$n_x$}\psfrag{ty}[l][l][.8]{$n_y$}\psfrag{tz}[l][l][.8]{$n_z$}
\centering
\includegraphics[width=9.5cm]{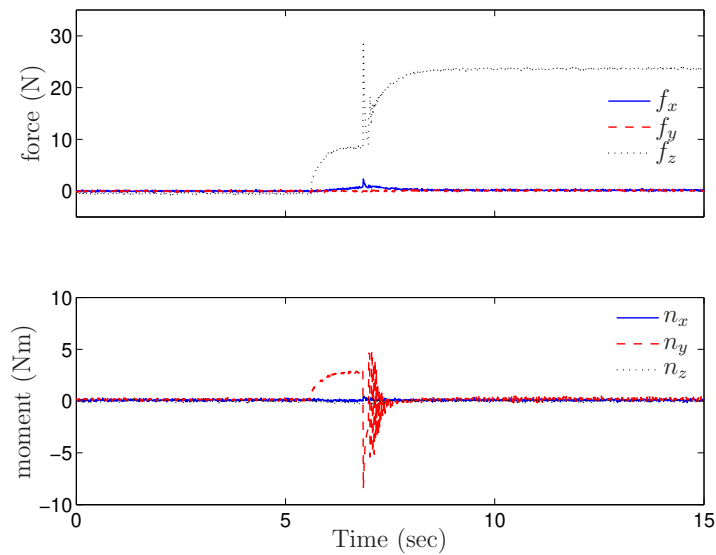}
\caption{Contact force and moment.}\label{fig:contact_force}
\end{figure}

\begin{figure}
\psfrag{vx}[l][l][.8]{$v_x$}\psfrag{vy}[l][l][.8]{$v_y$}\psfrag{vz}[l][l][.8]{$v_z$}
\psfrag{wx}[l][l][.8]{$\omega_x$}\psfrag{wy}[l][l][.8]{$\omega_y$}\psfrag{wz}[l][l][.8]{$\omega_z$}
\psfrag{v}[c][c][.8]{$\bm
v$~(m/s)}\psfrag{omeg}[c][c][.8]{$\bm\omega$~(rad/s)}
\centering\includegraphics[clip,width=9.5cm]{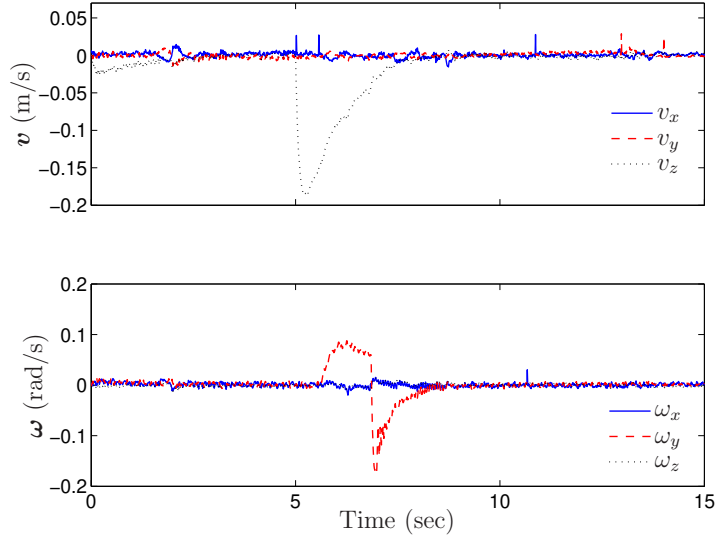}
\caption{Trajectories of the linear and angular
velocities.}\label{fig:contact_vel}
\end{figure}

In the second case study, the stability of the impedance control
system involving contact task is investigated. In this experiment,
the manipulator squeezes its payload against a flat horizontal
surface. The standard mass-damper-spring model is chosen as the
target impedance with parameters according to
\eqref{eq:target_inertia} and
\begin{align*}
\bm K_d &= \mbox{diag}\{470, 470, 470, 10, 18, 20 \}\\
\bm D_d &= \mbox{diag}\{600, 600, 600, 12, 20, 25 \}
\end{align*}
Note that the damping matrix is chosen so as to achieve a
well-damped behavior with damping ratio $\zeta=2$. Initially, the
manipulator holds the box just a few centimeters above the flat
surface. In order to established a contact force between the box and
the environment the manipulator is commanded to move along the
z-axis to a position obstructed by the flat surface, i.e.,  $\bm x_d
=\col{0, 0, x_{d_z}, 0, 0 ,0}$ and  $\dot{\bm x}_d = \ddot{\bm x}_d
= \bm 0$. Fig.~\ref{fig:contact_pos} illustrates the actual position
of the payload along the z-axis versus the desired one. At $t<5$s
when no contact is encountered, the controller regulates the robot
position to the desired one just above the surface. However, at
$t<5$s the robot is commanded to a position so that the box is
obstructed by the environment that leads to the contact  at $t=6$s.
Fig.~\ref{fig:contact_force} shows trajectories of the contact force
and moment, while Fig.~\ref{fig:contact_vel} shows trajectories of
the linear and angular velocities. The contact force along z-axis,
which stabilizes around $23$~N, and the steady-state position error,
$5$~cm, are closely related to the corresponding $470$~N/m stiffness
component of the enforced impedance. It is clear from
Fig.~\ref{fig:contact_force} that, in addition to the z-axis force,
significant external moment along y-axis is initially built up due
to the misalignment of the box surface with respect to the
environment. However, the impedance controller responds to the
moment and causes the box to slip in a way that eventually leads to
significant reduction of the built-up moment.


\section{Conclusions}

An impedance controller for manipulators carrying a rigid-body
payload has been developed that does not rely on any acceleration
measurements in order to compensate for the inertial forces of the
payload. The stability and convergence of the impedance controller
have been analytically investigated. The results showed that the
control input remains bounded  provided that the desired inertia was
selected to be different from that of the payload.  The impedance
controller was further developed with utilizing an inner/outer loop
approach. This allows specifying the desired impedance model in a
general form. The stability analysis revealed that such an impedance
control approach has its own limitation; that is to maintain the
stability of the closed-loop system, the ratio of payload inertia to
the manipulator Cartesian inertia must be lower than a ceratin value
(which depends on the feedback gains). It should be noted that this
condition is in addition to the well-known condition on the contact
stability, which imposes a lower bound limit on the magnitude of the
ratio of the desired inertia matrix to the actual effective inertia
matrix when the impedance control is implemented digitally or the
actuators are of finite bandwidth. Furthermore, robust control
theory was employed to modify the proposed impedance controller so
that it could compensate for robot modeling uncertainty.
Subsequently, the conditions for achieving robust impedance control
in the face of modeling uncertainty have been derived. The
conditions for this, however, were found to be more stringent than
those of the conventional robust position control of manipulators.
Experimental results have shown that the proposed impedance
controller enabled a manipulator carrying a payload with a
non-negligible mass to establish the desired impedance
characteristic even though no acceleration measurements were used.

\appendix
\section*{Appendix A: Manipulator \& its Payload} \label{apdx:cartesian_dynamics}
Dynamics equations of the manipulator in joint space are given by
\begin{equation} \label{eq:Mrddotq}
{\bm M}'_m \ddot{\bm q} + \bm {h}'_m(\bm q, \dot{\bm  q}) = \bm\tau
+ \bm J^T \bm f_s
\end{equation}
Substituting the joint acceleration from $\ddot{\bm q}=\bm J^{-1}
\ddot{\bm x} - \bm J^{-1} \dot{\bm J}\dot{\bm q}$ into
\eqref{eq:Mrddotq} and then multiplying the resultant equation by
$\bm J^{-T}$ yields \eqref{eq:manipulator_imp}, in which
\begin{align*}
\bm M_m & \triangleq \bm J^{-T} {\bm M}'_m \bm J^{-1} \\
\bm h_m &\triangleq \bm J^{-T} {\bm h}'_m - \bm M_m \dot{\bm J}
\dot{\bm q}
\end{align*}

\section*{Appendix B: Stability of the Perturbed System} \label{apdx:Lyapunov}
Since $\bm A$ is Hurwitz, there exists a Lyapunov function
\begin{equation} \label{eq:Lyapunov}
V( \bm z) = \bm z^T \bm P \bm z
\end{equation}
with $\bm P> 0 $ satisfying
\begin{equation} \label{eq:PA}
\bm P \bm A + \bm A^T \bm P = -\bm 1.
\end{equation}
The derivative of $V(\bm z)$ along trajectories of the perturbed
system \eqref{eq:perturbed} satisfies
\begin{equation}
\dot V \leq \Big(-1  + 2 \norm{\bm G } \lambda_{\rm max}(\bm P)
\norm{\bm\Delta_m} \Big) \norm{\bm z}^2.
\end{equation}
Therefore, according to the stability theorem of perturbed systems
\cite[p. 206]{Khalil-1992}, the origin of \eqref{eq:perturbed} is
globally exponentially stable if
\begin{equation} \label{eq:cond_Deltam}
\norm{\bm\Delta_m} \leq \kappa(\bm G_p, \bm G_d) \triangleq
\frac{1}{2 \norm{\bm G} \lambda_{\rm max}(\bm P)}.
\end{equation}

\section*{Appendix C: Uncertainty Bound} \label{apdx:rho}
Assuming \eqref{eq:rho_eta} and the following
\begin{equation} \notag
\|\bm f_{\rm ext}\| \leq c_f, \quad  \| \bm h_d \| \leq c_d, \quad
\| \tilde{\bm h}_t \| \leq c_t,
\end{equation}
and choosing $\rho$ such that
\begin{equation}
\rho \leq \frac{c_f + c_d}{1+ \alpha}  \| \bm\Delta \tilde{\bm M}_t
\|  + \frac{c_t}{1+ \alpha} \| \bm\Delta \bm M_t^{-1} \|  +
\frac{\alpha}{1 + \alpha} \| \bm G \| \| \bm z \|
\end{equation}
will satisfy \eqref{eq:alpha}.
\section*{Appendix D: External Force Estimate} \label{apdx:external_force}
The acceleration term can obtained from \eqref{eq:manipulator_imp}
as
\begin{equation} \label{eq:ddotx}
\ddot{\bm x}= \bm M_m^{-1}\big( -\bm h_m + \bm u + \bm f_s \big).
\end{equation}
Substituting \eqref{eq:ddotx} into \eqref{eq:payload_imp} yields
\begin{equation}
\bm f_{\rm ext} = \bm h_p - \bm M_p \bm M_m^{-1} \bm h_m + \bm M_p
\bm M_m^{-1} \bm u + \bm( \bm 1 + \bm M_p \bm M_m^{-1} \big) \bm
f_s,
\end{equation}
which can be used to compute the external force as a function of
force sensor output, $\bm f_s$, and the control input, $\bm u$,
issued by the impedance controller.

\bibliographystyle{IEEEtran}

\end{document}